# Multimodal Sensor Dataset for Monitoring Older Adults Post Lower-Limb Fractures in Community Settings

Ali Abedi[1*], Charlene H. Chu[1,2], and Shehroz S. Khan[1,3]

[*]Correspondence: ali.abedi@uhn.ca
[1] KITE Research Institute, Toronto Rehabilitation Institute, University Health Network, Toronto, Canada
[2] Lawrence Bloomberg Faculty of Nursing, University of Toronto, Toronto, Canada
[3] College of Engineering and Technology, American University of the Middle East, Kuwait

**Abstract**
Lower-Limb Fractures (LLF) are a major health concern for older adults, often leading to reduced mobility and prolonged recovery, potentially impairing daily activities and independence. During recovery, older adults frequently face social isolation and functional decline, complicating rehabilitation and adversely affecting physical and mental health. Multi-modal sensor platforms that continuously collect data and analyze it using machine-learning algorithms can remotely monitor this population and infer health outcomes. They can also alert clinicians to individuals at risk of isolation and decline. This paper presents a new publicly available multi-modal sensor dataset, MAISON-LLF, collected from older adults recovering from LLF in community settings. The dataset includes data from smartphone and smartwatch sensors, motion detectors, sleep-tracking mattresses, and clinical questionnaires on isolation and decline. The dataset was collected from ten older adults living alone at home for eight weeks each, totaling 560 days of 24-hour sensor data. For technical validation, supervised machine-learning and deep-learning models were developed using the sensor and clinical questionnaire data, providing a foundational comparison for the research community.

## 1. Introduction

Lower-Limb Fractures (LLF) are a significant health issue, particularly among older adults [1], [2]. As the global population ages, the incidence of LLF is projected to rise, with prevalence rates for hip and LLF expected to double by 2050 [1], [3]. Recovery from these fractures often involves prolonged mobility limitations, impacting activities of daily living and quality of life [4]. Post-surgical complications, including reduced muscle strength and age-related conditions such as osteoporosis, exacerbate recovery challenges [5]. Psychological distress and cognitive decline further hinder recovery, leaving only 40–60% of hip fracture patients regaining pre-fracture mobility [6].

Social isolation is prevalent among older adults recovering from fractures, particularly those living alone [7]. Defined as minimal social contact, it is closely linked to functional decline, creating a cycle that negatively impacts independence and well-being [8], [9]. Addressing social isolation and functional decline is critical to improving recovery outcomes [10]. Research highlights the importance of early identification of older adults following LLF who are at risk of these issues, as preventive programs can help reduce readmissions [11] and further enhance recovery [10]. However, standard care often lacks routine assessments for social isolation and functional decline, two critical determinants of recovery and long-term well-being, particularly in orthopedic and geriatric populations [12], [13]. Social Isolation assessments are not typically part of primary care, as clinicians often rely on patients to self-report their experiences of isolation. However, due to social stigma, many individuals may avoid reporting feelings of loneliness or disconnection [14]. Similarly, functional decline can only be assessed during clinic visits, creating a significant gap in monitoring patients who may be experiencing gradual but impactful physical deterioration at home [15]. This reliance on sporadic, patient-initiated reporting represents a missed opportunity for both patients and clinicians, as these issues often remain undetected until they contribute to significant health setbacks.

To address these challenges, we emphasize the need for automatic, remote assessment tools capable of continuously monitoring social isolation and functional decline in a non-invasive manner. Recent advancements in digital health technologies provide promising avenues for passive data collection [16], [17], leveraging sensors and Artificial Intelligence (AI) to collect longitudinal data on mobility and social interactions. These tools allow for early detection of risk factors, enabling timely, targeted interventions. Social reintegration initiatives informed by such tools can help mitigate isolation, improving mental health and overall well-being [18]. Similarly, accurate, real-time mobility and functional assessments can optimize physical therapy regimens, foster self-sufficiency, and improve quality of life [19].

To this end, we introduce the MAISON-LLF dataset containing multimodal sensor data and clinical questionnaire data on social isolation and functional decline. Preliminary investigations on a smaller subset of this dataset [9] reveal strong correlations between the sensor data and the corresponding questionnaire data, highlighting the potential of AI-driven predictive modeling to enable personalized recovery strategies. We conduct extensive validation experiments to establish the dataset's reliability for both technical and clinical research purposes. These experiments demonstrate how MAISON-LLF can support AI-driven predictive modeling, laying the groundwork for a comprehensive, data-driven approach to orthopedic and geriatric care, ultimately improving recovery outcomes and enhancing patients' quality of life.

## 1.1. Literature Review

Previous research on leveraging sensors and AI to detect either social isolation or functional decline in older adults remains limited. Khan et al. [12] conducted a literature review on sensor-based approaches for assessing social isolation among community-dwelling older adults and categorized the previous works focusing on physiological, phone communication, and mobility data. Among these, the only AI-driven approach included in their review was by Martinez et al. [20], who utilized communication and home mobility data with a Decision Tree (DT) model to classify social isolation levels. Jiang et al. [21] and McCrory et al. [22] identified links between social isolation, sleep metrics, and resting heart rate. Goonawardene et al. [23] used motion and door sensors to correlate going-out behavior, daytime naps, and time in the living room with isolation. Prenkaj et al. [24] employed smartwatch-collected data (location, heart rate, sleep, mood) and self-supervised learning to detect anomalies suggesting early signs of isolation. A systematic review on home-based digital interventions for older adults post-hip fracture surgery [25] identified only one sensor study in nursing homes [26], where wearable accelerometers recognized daily activities and improved outcomes. Kraaijkamp et al. [27] utilized accelerometers to measure physical activity in older adults recovering from hip fractures, identifying consistently low activity levels and extended periods of sedentary behavior across all participants during a seven-day data collection period. Braun et al. [28] combined step counts and demographic data using logistic regression, achieving over 80% accuracy in predicting recovery likelihood in orthopedic trauma patients.

Although no known data specifically addresses the assessment of social isolation and functional decline in older adults, some multimodal sensor datasets exist for other populations; however, each comes with its own limitations. For instance, Technology Integrated Health Management (TIHM) [29] recorded sleep, motion, skin temperature, blood pressure, heart rate, muscle mass, weight, and body water from people with dementia in their homes over 50 days. The data was not collected continuously; for example, heart rate measurements were limited to one value per day, which participants were required to record manually. Events were detected by thresholding and machine-learning without human-annotated ground truth. GSTRIDE [30] collected inertial data from frail older adults during a single 30-minute walking test in nursing homes, lacking multiple modalities and longitudinal follow-up. Another dataset [31] from 20 people with dementia included accelerometer, blood volume pulse, electrodermal activity, and skin temperature with human-annotated agitation. However, it lacked home-based data.

## 1.2. Gaps in the Literature and Study Aim

A significant gap persists in detecting both social isolation and functional decline among older adults recovering from LLF, particularly within their home environments. To the best of our knowledge, no studies have examined these two factors simultaneously under such conditions. Existing research often relies on a single data modality, such as indoor motion [32] or step count [33], despite calls for comprehensive, long-term, multimodal sensor data [12], [34]. Moreover, most studies focus on identifying correlations rather than applying predictive models [9], [21], [22], [23]. While correlation analysis is a valuable first step to demonstrate the usefulness of sensors in assessment, predictive modeling is essential for making automated decisions that support clinical care and enhance the overall quality of life for recovering patients. To date, no publicly available multimodal dataset exists for the LLF population living in the community, limiting the development of robust predictive modeling approaches. To address these gaps, this study aimed to collect a novel multimodal sensor dataset by monitoring ten participants over eight weeks each in community settings using a multimodal sensor platform. The dataset is publicly released to advance predictive analytics for social isolation and functional decline. Using this dataset, the study developed and evaluated machine-learning and deep-learning models to detect both conditions in this population.

## 2. Methods

This section describes the multimodal sensor platform developed and deployed in the homes of older adults recovering from LLF, the data collected, and the preprocessing steps undertaken to prepare these data for machine-learning model development for predicting social isolation and functional decline.

### 2.1. MAISON: Multimodal AI-based Sensor platform for Older iNdividuals

Multimodal AI-Based Sensor Platform for Older iNdividuals (MAISON) [35] was developed to integrate in-home Internet-of-Things and remote monitoring technologies, enabling the continuous acquisition of routine physiological, sleep, location, movement, and ambient data. Figure 1 illustrates the block diagram of MAISON [35], which comprises the MAISON smartphone (phone) application (app), the MAISON smartwatch (watch) app, external non-wearable sensors with their respective clouds, and the MAISON central cloud.

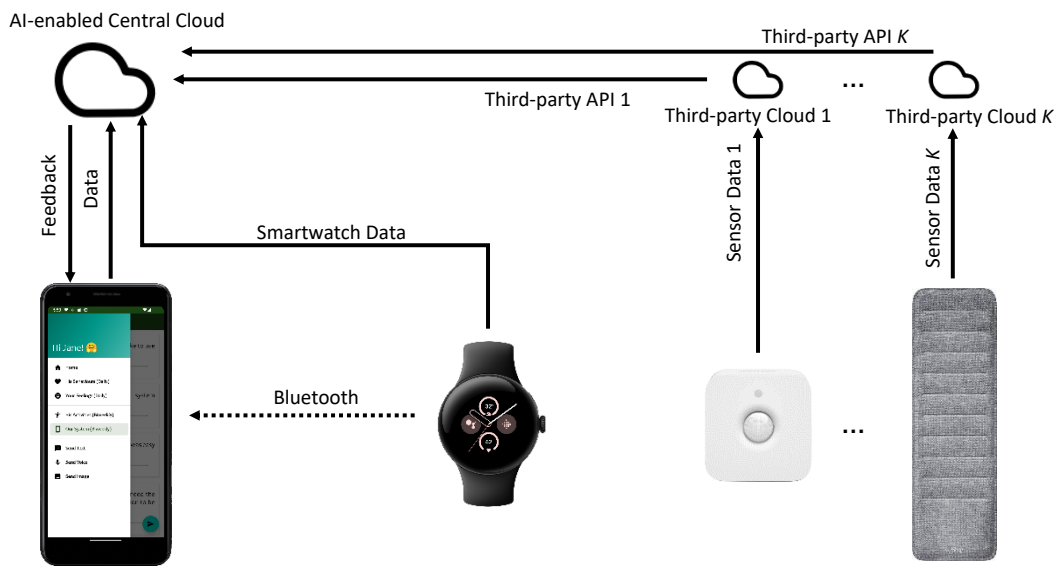

**Figure 1.** The block diagram of MAISON [35], as described in subsection 2.1.

The phone app is designed to collect and store data from the phone's built-in sensors [36], collect questionnaire responses, connect with the watch to receive and store the watch's built-in sensors data [36], and transfer data to the cloud. It features a user-friendly interface for older adult and patient participants. Participants can switch data collection on and off. The app automatically resumes data collection after recharging or rebooting to enhance ease of use.

MAISON can leverage any Android phone sensor, such as the Global Positioning System (GPS) and accelerometer [36], for data collection and cloud storage. The sampling frequency, duration of data collection, and specific questionnaires and forms within the phone app are customizable according to study requirements. To conserve battery life, MAISON uses geofencing to track GPS, activating GPS data collection only when participants move beyond a defined perimeter of their home [37]. All data is stored locally on the device in real-time and is periodically transferred to the cloud, which is hosted on Google Cloud Platform [38] and dedicated exclusively to MAISON.

In addition to the phone, the watch's built-in sensors collect physiological and activity data, such as heart rate, accelerometer, and step count, transmitting data directly to the cloud. When offline, data is transferred to the phone via Bluetooth. MAISON employs Google Android for phones and Google Wear Operating System (OS) for watches, ensuring full compatibility and continuous connectivity [36].

MAISON can integrate any external commercial sensors, provided the manufacturer offers an Application Programming Interface (API) to access the data collected on their cloud and transfer it to the MAISON central cloud. Using manufacturer-provided APIs and secure authentication, data from these devices is automatically retrieved from their respective clouds and stored on the MAISON central cloud.

## 2.2. MAISON-LLF: MAISON-Lower Limp Fracture Dataset

This paper presents the MAISON-LLF dataset, distinguished by the following unique characteristics: (1) the dataset contains data collected from older adults living alone in the community after LLF surgery, (2) It features multi-modal sensor data that provides complementary and supplementary information about participants, (3) It includes continuous sequential and longitudinal data collected over multiple weeks post-fracture, (4) and it incorporates gold-standard ground-truth data from clinically validated questionnaires. Together, these features make the dataset well-suited for developing supervised and unsupervised machine-learning models for predicting different health outcomes.

### 2.2.1. Participants

Older adults who experienced LLF were directly recruited from the Toronto Rehabilitation Institute, with all data collection conducted within the Greater Toronto Area, Canada. The study received approval from the University Health Network's Research Ethics Board (study ID 20-5113), and informed consent was obtained from all participants, allowing full public release of their de-identified data.

The inclusion criteria required participants to be 60 years or older, have undergone surgical repair for hip, femur, or pelvis fractures, or have had a hip replacement. Participants also needed to have a Wi-Fi connection at home. They should not have cognitive impairment, indicated by a Mini-Mental State Examination (MMSE) score of 24 or higher [39]. Recruitment and enrollment began prior to participant discharge, enabling data collection to start immediately within the first few days following discharge.

### 2.2.2. Multi-Modal Sensor Data Collection

MAISON was employed to collect data from participants within the community over a period of eight weeks immediately following discharge from hospital. This duration was selected based on previous research indicating that most physical recovery occurs in the first few months after discharge [40]. MAISON for this study is comprised of
- a Motorola Moto G7 Android phone [41],
- a Mobvoi TicWatch Pro Wear OS watch [42],
- a Withings Sleep sleep-tracking mat [43], and
- an Insteon Motion Sensor II motion detection sensor [44].

The MAISON phone and watch apps automatically collected data from participants continuously both inside and outside their homes, and transferred it to the cloud every 30 minutes. GPS location data were continuously collected via the phone when participants were detected to be outside the home. Through the watch, acceleration data were collected every second, heart rate every 30 minutes, and step count continuously. The watch required one and a half hours of charging every twelve hours, while the phone needed two hours every other day. Participants only need to recharge the watch and phone, wear the watch, and carry the phone when going outside; no further actions are required for data collection or cloud transfer. The data collection periods were carefully planned to collect the most useful information while minimizing the need for recharging since data collection stops when the device is being charged.

Sleep data were collected during each sleep session using the sleep-tracking mat positioned beneath the bed mattress and connected to a wall outlet. These data were subsequently transferred to the cloud after each sleep session. Additionally, motion data were continuously collected via the motion detection sensor located in the living room and connected to a wall outlet, facilitating real-time data transfer to the cloud. Refer to Table 1 for detailed information on the collected sensor data modalities.

### 2.2.3. Demographic and Gold-standard Clinical Data Collection

Demographic information, including sex, age, LLF type, relationship status, education level, work status, and ethnicity, was collected from participants at the start of the study. Every two weeks, validated clinical assessments were administered by a trained research assistant via Microsoft Teams video calls with participants at home. These clinical assessments, serving as gold-standard data, included the Social Isolation Scale (SIS) [45] for assessing social interactions, the Oxford Hip Score (OHS) [46], the Oxford Knee Score (OKS) [47], the Timed Up and Go (TUG) test [48], and the 30-second chair stand test [49] for evaluating functional health.

**Table 1.** Data collection devices, data modalities collected through the devices, and the period of data collection (An event-based data is collected when an event occurs, such as a participant taking a step) along with the clinical gold-standard data.

| Device | Sensor Data Modality / Clinical Gold-standard Data | Period/Event-based |
|---|---|---|
| Smartwatch | - Acceleration (x, y, and z coordinates in m/s²)<br>- Heartrate value (measured in beats per minute)<br>- Step (a detected step) | - 1 second<br>- 30 minutes<br>- Event-based |
| Smartphone | - GPS location (latitude and longitude coordinates) | - 10 seconds |
| Motion Sensor | - Motion (a detected motion) | - Event-based |
| Sleep-tracking Mat | - Total, deep, light, and rapid-eye-movement sleep duration, snoring duration, duration to sleep and to wake up (in hours), heart rate during sleep (beats per minute), and wake-up count (in counts) | - 24 hours (once at each sleep session) |
| Questionnaire / Test | - Social Isolation Scale (6 items questionnaire, total 6–30)<br>- Oxford Hip Score (12-item questionnaire, total 0–48)<br>- Oxford Knee Score (12-item questionnaire, total 0–48)<br>- Timed Up and Go (physical test in seconds)<br>- 30-second chair stand (physical test in counts) | - Every other week |

The 6-item SIS questionnaire [45] assesses specific objective and subjective constructs of social interactions and is designed for older adults. For example, one question in SIS asks, "How many of your family, friends, and neighbors do you see face-to-face at least once?" [45]. Each item is rated on a 1 to 5 Likert scale, totaling 6 to 30 points, with higher scores indicating more social interaction and lower social isolation. The 12-item OHS [46] and 12-item OKS [47] questionnaires are standard clinical tools that are routinely used to assess the outcomes of lower limb fracture surgery, focusing specifically on a patient's physical function and pain level. For instance, a question in OHS is, "Have you been limping when walking because of your hip?" [46]. Each item is rated on a 0-4 Likert scale, with total scores ranging from 0 to 48; higher scores indicate better physical functioning and lower pain levels. Refer to Table 1 for a detailed summary of the collected clinical data.

### 2.2.4. Preprocessing and Feature Extraction

A key part of preprocessing data for feature extraction, analysis, and predictive modeling involves managing the variations in collection periods between different sensor data and gold-standard questionnaire data. For example, acceleration data was collected every second, heart rate data every 30 minutes, and step count continuously. To accomplish this task, the data was aggregated and processed to extract features representing daily metrics, as detailed in Table 2. For example, the feature 'step-count' represents the total number of steps taken over a 24-hour period, from midnight to the following midnight. Alternatively, a similar set of features can be extracted to represent hourly metrics, e.g., 'step-count' representing the total number of steps recorded within a given hour.

The collected dataset exhibited an average of 6.7% missing data at the daily feature level, and imputation was conducted at this level. For days with a missing data modality, the features were imputed using the average values of that feature from the same participant on other days within the same two-week period of biweekly gold-standard data collection.

The clinical questionnaire data were designed to assess participants' social interaction and physical functioning over weeks and were collected accordingly, while sensor data was collected continuously. To address the sparsity of these clinical outcome values, three approaches for outcome assignment and corresponding data sample creation were developed, as shown in Figure 2. The 14 days preceding the day on which the clinical outcomes were collected were utilized in three ways:

- Treated as 1 fourteen-day data sample, with the clinical outcomes assigned to it.
- Divided into 2 seven-day data samples, with the same clinical outcomes assigned to both seven-day data samples.
- Treated as 14 individual daily data samples, with the same clinical outcomes assigned to all days.

The first and second approaches provide more detailed information in each data sample by capturing changes in data values across consecutive days. However, compared to the third approach, they result in smaller numbers of data samples.

## 3. Data Records

This section provides an overview of the participants whose data are included in the MAISON-LLF dataset. It then details the repository where the dataset is stored, along with its folder structure and data files. Finally, it presents a brief summary of the dataset's key statistics.

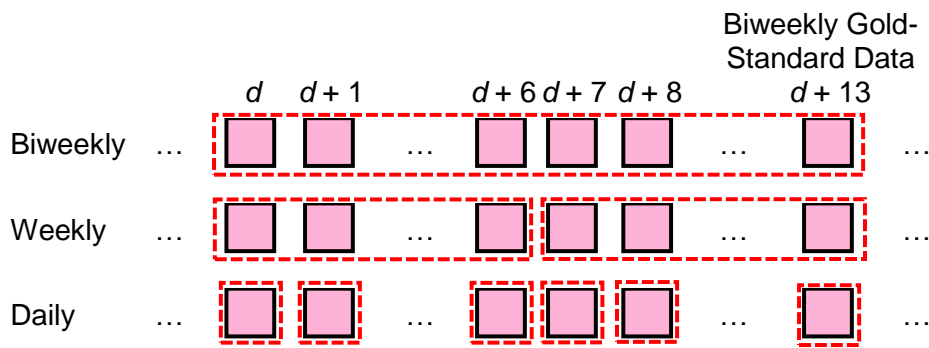

**Figure 2.** Three approaches for outcome assignment and data sample creation: biweekly, weekly, and daily, resulting in 1, 2, and 14 data samples per gold-standard data, respectively. $d$, $d + 1$, …, $d + 13$ are 14 days before the day a biweekly gold-standard data was collected.

**Table 2.** The data modalities, feature names, and their description.

| data modality-[1] | feature [1] | Description |
|---|---|---|
| acceleration- | count, entropy, kurtosis, mean, skew, std, and sum | The total count, entropy, kurtosis, mean, skewness, standard deviation, and sum of acceleration data in a day. |
| heartrate- | max, mean, min, and std | The maximum, mean, minimum, and standard deviation of heartrate in a day. |
| motion (step)- | count | The total count of motions (steps) per day. |
| | max | The maximum number of motions (steps) in hours of a day. |
| | max-timestamp | The timestamp (hour of day) in which there has been maximum number of motions (steps) in a day. |
| | mean | The average number of motions (steps) in hours of a day with at least one motion (step). |
| | ratio | The ratio of the number of hours with at least a motion (step) to the number of hours without any motion (step) in a day. |
| position- | count | The total count of position data in a day. |
| | duration | The duration (in hours) of being outside the home in a day. |
| | travelled-distance | The total distance (in kilometers) traveled outside the home in a day. |
| sleep- | deep, light, rem, snoring, and total | The duration (in hours) of deep, light, rapid-eye-movement, snoring, and total sleep. |
| | duration-to-sleep and duration-to-wakeup | The duration (in hours) to sleep and to wake up. |
| | heartrate-max, heartrate-mean, and heartrate-min | The maximum, mean, and minimum heartrate during sleep. |
| | wakeup-count | The count of wakeups during sleep time. |
| demographic- | age, education, ethnicity, fracture, relationship, sex, and work | The age, education level, ethnicity, lower-limb fracture type, relationship status, sex, and work status of the participant. |

[1] Feature naming convention in the dataset: 'data modality' + '-' + 'feature'

## 3.1. Participants

The MAISON-LLF dataset was collected from 10 older adult participants living alone in the community following LLF. The data collection described in this paper took place between March 2022 and December 2024. Each participant contributed data for over 8 weeks (56 days), beginning from their first week post discharge. This resulted in a total of 560 days of continuous multimodal sensor data, complemented by biweekly clinical questionnaire data. Table 3 presents the demographic information of the participants. Ongoing data collection by the research team aims to increase the number of participants to 20.

**Table 3.** Demographic information of participants included in the MAISON-LLF dataset.

| #  | Sex    | Age | Fracture Type    | Relationship | Education     | Work      | Ethnicity |
|----|--------|-----|------------------|--------------|---------------|-----------|-----------|
| 1  | Female | 74  | Hip Fracture     | Widowed      | Undergraduate | Retired   | Black     |
| 2  | Female | 66  | Hip Fracture     | Separated    | Undergraduate | Part-Time | White     |
| 3  | Female | 70  | Pelvis Fracture  | Single       | Secondary     | Retired   | White     |
| 4  | Female | 60  | Femur Fracture   | Divorced     | Secondary     | Retired   | White     |
| 5  | Male   | 75  | Hip Replacement  | Single       | Doctorate     | Part-Time | White     |
| 6  | Female | 85  | Pelvis Fracture  | Single       | Undergraduate | Part-Time | White     |
| 7  | Female | 78  | Hip Replacement  | Divorced     | Graduate      | Retired   | White     |
| 8  | Female | 89  | Hip Replacement  | Single       | Secondary     | Retired   | White     |
| 9  | Female | 74  | Hip Fracture     | Divorced     | Doctorate     | Retired   | White     |
| 10 | Male   | 77  | Hip Replacement  | Widower      | Secondary     | Retired   | White     |

### 3.2. Repository, Folder Structure, and Data Files

The MAISON-LLF dataset is available on Zenodo [50], a platform that enables sharing and preserving digital research objects, including datasets, publications, and software. The MAISON-LLF dataset is structured in a directory tree illustrated in Figure 3 and described below.

In 'sensor-data' folder, the dataset includes 60 CSV files containing data from six sensor types for 10 participants. Each CSV file includes a 'timestamp' column indicating the date and time of the recorded sensor data, accurate to milliseconds ('yyyy-MM-dd HH:mm:ss.SSS'), along with the corresponding sensor measurements as described in Table 1. For instance, the 'acceleration-data.csv' files include four columns: timestamp, and x, y, and z coordinates of acceleration, while the 'heartrate-data.csv' files contain two columns: timestamp and heart rate value.

The dataset also includes 70 CSV files, in 'features' folder, containing daily features extracted from the sensor data, along with clinical questionnaire data and physical test results. Each feature CSV file includes a timestamp column representing the date ('yyyy-MM-dd') of the sensor data from which the daily features were extracted, alongside the corresponding sensor features as detailed in Table 2. For example, the 'acceleration-features.csv' files contain eight columns: timestamp and the seven acceleration features described in Table 2 and the 'heartrate-features.csv' files include five columns: timestamp and the four heart rate features outlined in Table 2. Additionally, the 'clinical.csv' files provide values for individual items of the SIS ('sis-01' to 'sis-06'), OHS ('ohs-01' to 'ohs-12'), and OKS ('oks-01' to 'oks-12') questionnaires, along with their final scores ('sis', 'ohs', and 'oks'). These files also include results for the TUG and 30-second chair stand tests. Each participant has four sets of clinical data, with each set sharing the same 'timestamp' corresponding to the date ('yyyy-MM-dd') on which the clinical data were collected.

To provide a comprehensive overview of the dataset, the 'all-features.csv' and 'all-features-imputed.csv' files in 'dataset' folder combine all daily features, clinical data, and demographic information into single CSV files, representing the data before and after missing value imputation (as explained in subsection 2.2.4). Additionally, the Python PyTorch files are structured datasets designed to facilitate supervised and unsupervised machine learning model development for estimating clinical outcomes.

'dataset-daily.pt' in 'dataset' folder contains a NumPy array with dimensions num_days × num_features, representing the daily features for all 10 participants. Alongside this array, it includes a num_days IDs array that maps each day to a participant (IDs 1 to 10). Additionally, the file contains three separate num_days arrays for SIS, OHS, and OKS scores, each assigned to the corresponding days in the daily features array.

'dataset-weekly.pt' in 'dataset' folder provides an array with dimensions num_weeks × 7 × num_features, which includes the weekly sequential features for all participants. This file also includes a num_weeks IDs array to identify the participant (1 to 10) associated with each week in the samples array. Similar to the daily dataset, it contains three separate num_weeks arrays for the SIS, OHS, and OKS scores, each assigned to the respective weeks in the weekly features array.

'dataset-biweekly.pt' in 'dataset' folder provides an array with dimensions num_biweeks × 14 × num_features, which includes the biweekly sequential features for all participants. This file also includes a num_biweeks IDs array to identify the participant (1 to 10) associated with each biweekly period in the samples array. Similar to the daily dataset, it contains three separate num_biweeks arrays for the SIS, OHS, and OKS scores, each assigned to the respective biweekly periods in the biweekly features array.

Across these files, the values of num_days, num_weeks, num_biweeks, and num_features are 560, 80, 40, and 35, respectively. 'demographic.csv' contains the demographic information of the 10 participants (Table 3).

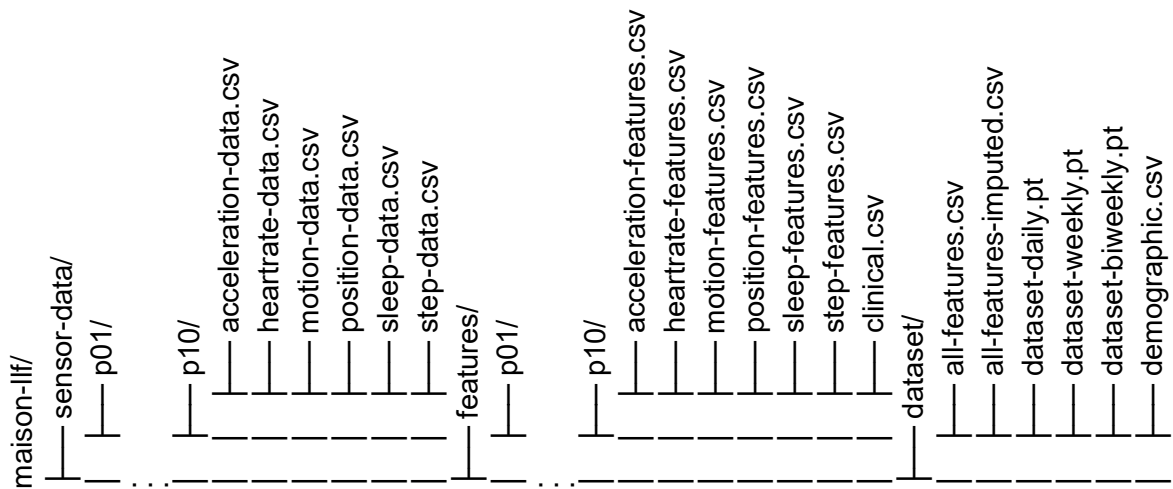

**Figure 3.** Directory tree of the MAISON-LLF dataset, including the sensor data, features extracted from the sensor data (the file structure is consistent across participants, from p01 to p10), and the constructed dataset in different settings.

### 3.3. Key Statistics

Figure 4 provides an example visualization of the daily features extracted from six data modalities for one participant over 8 weeks of data collection (see Table 2 for feature descriptions). These modalities capture complementary information: motion sensors track movements indoors, GPS tracks location outdoors, watches monitor steps, heart rate, and acceleration when participants are awake, and sleep mats measure sleep characteristics during sleep.

Figure 5 illustrates the distribution of clinical questionnaire data collected across participants, providing insights into the variability and range of these measures within the dataset. Dayyani et al. [9] reported moderate to strong Spearman's rank correlations between daily features and clinical data in a subset of the MAISON-LLF dataset, highlighting the potential of multimodal sensor data to capture social interaction and physical health outcomes.

## 4. Technical Validation

This section validates the technical capability of the MAISON-LLF dataset in capturing health outcomes in older adults post-LLF living alone in the community. The features extracted from the multimodal sensor data and selected through data-driven feature-selection algorithms were used as input for machine-learning and deep-learning models trained to estimate health outcomes. For validation, SIS and OHS were chosen as ground-truth labels, while similar experiments can be extended to other outcomes.

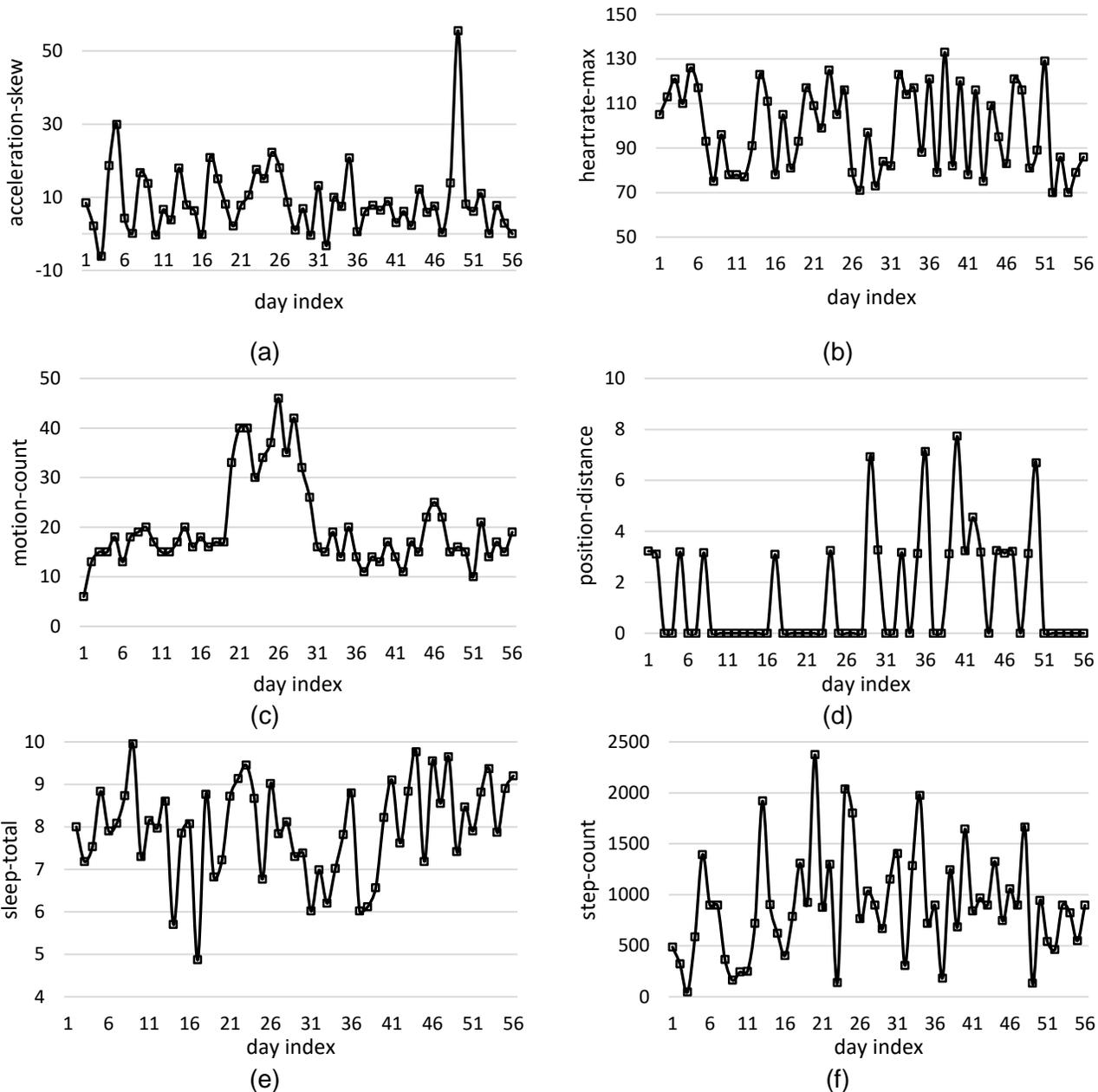

**Figure 4.** An example visualization of the daily features for one participant over 8 weeks of data: (a) acceleration-skew, (b) heartrate-max, (c) motion-count, (d) position-travelled-distance, (e) sleep-total, and (f) step-count (see Table 2 for feature descriptions).

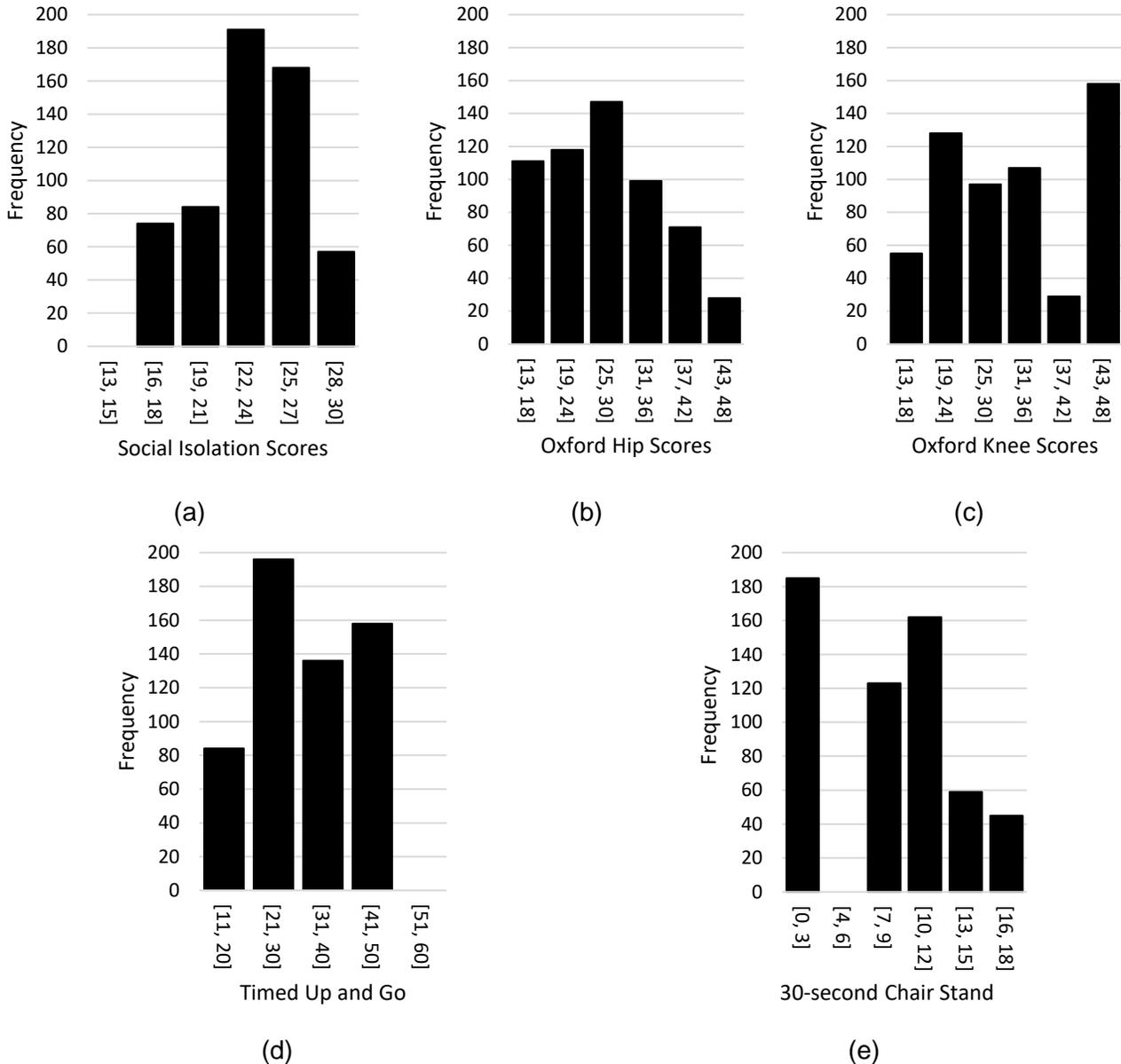

**Figure 5.** Histograms of clinical questionnaire data: (a) Social Isolation Scale, (b) Oxford Hip Score, (c) Oxford Knee Score, (d) Timed Up and Go test, and (e) 30-second Chair Stand test.

### 4.1. Feature Selection

The Recursive Feature Elimination (RFE) approach [51], [52] was used for feature selection that recursively fits a core predictive model and removes the least important features based on their importance scores until the desired number of features is reached.

Using the daily features described in Table 2 and applying the daily outcome assignment approach outlined in Figure 2, Categorical Boosting (CatBoost) [53] served as the core predictive model for RFE with SIS and OHS as the ground-truth outcomes. Figure 6 (a) and (b) present the SHapley Additive exPlanations (SHAP) values [54] for the top 24 features for SIS and OHS selected through RFE conducted on the dataset using 5-fold Cross Validation (CV) with CatBoost as the core classifier [51], [52]. CatBoost was utilized for visualization in this subsection, while various predictive models, each with its specific RFE-based feature selection, were evaluated as detailed in subsection 4.2. SHAP values quantify each feature's contribution to the deviation of the model's output from the baseline or average model output [54]. As shown in Figure 6, features from all data modalities were included among the top 24 selected by RFE, demonstrating the value of using multiple sensor modalities for precise health outcome assessment. As an example of interpreting SHAP values, the values for 'sleep-deep' show that higher values of this feature positively contributed to the model's SIS estimation. This implies that longer deep sleep resulted in higher SIS, indicating greater social interactions.

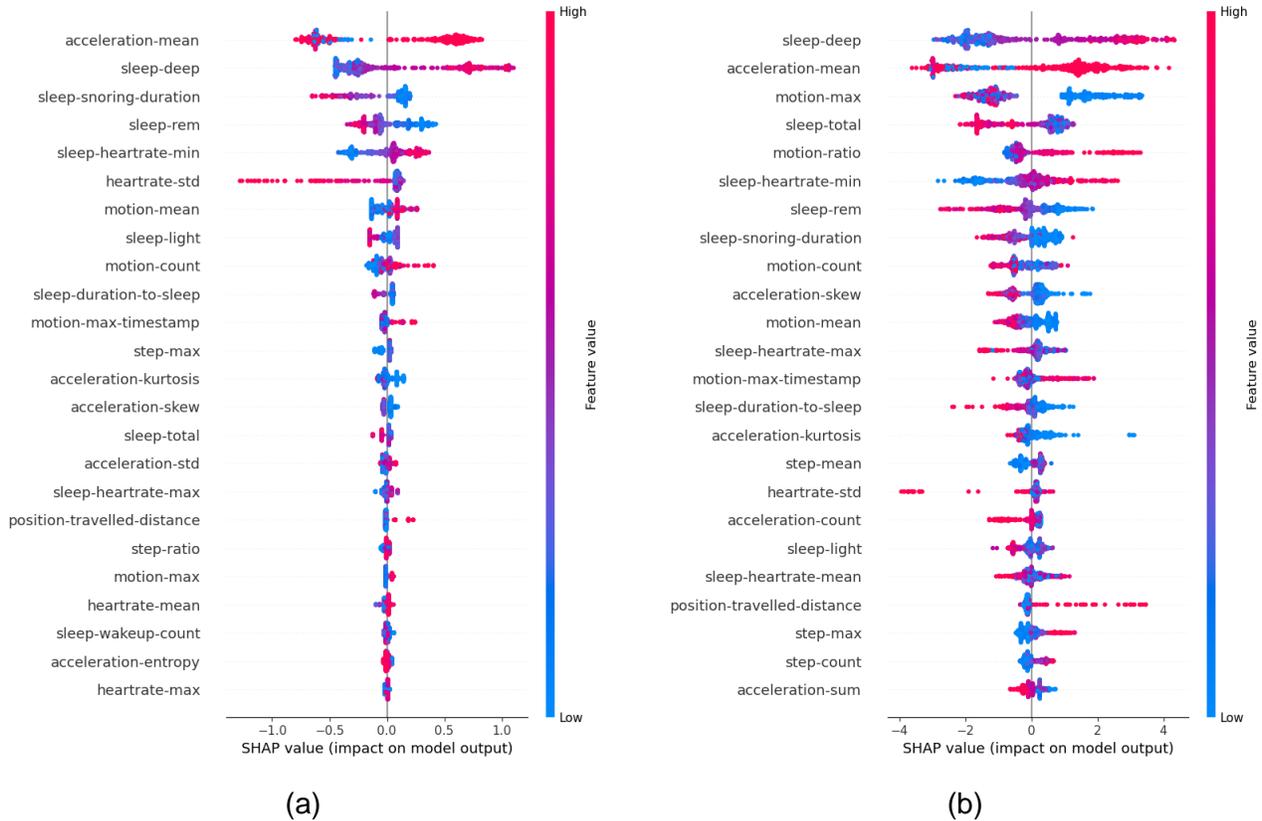

(a) (b)

**Figure 6.** Visualization of SHapley Additive exPlanations (SHAP) values, i.e., impact on Categorical Boosting (CatBoost) model output for regression of (a) Social Isolation Scale (SIS) and (b) Oxford Hip Score (OHS).

### 4.2. Predictive Modeling

The machine-learning and deep-learning models were trained in a supervised setting, where features were the input and SIS and OHS were the output to solve regression problems. For the daily outcome assignment approach, described in Figure 2, non-sequential models were used, including, Support Vector Regressor (SVR), DT [55], Light Gradient-Boosting Machine (LightGBM) [56], and CatBoost [53].

For the weekly and biweekly outcome assignment approach, models capable of analyzing sequential data were applied, including Long Short-Term Memory (LSTM), Temporal Convolutional Network (TCN), Bag of Words (BoW) [57], and ROCKET [58]. The evaluation metrics were Mean Absolute Error (MAE) and coefficient of determination (R2) [51], [52] between the outputs of predictive models and the ground-truth outcomes. The feature selection, model training, and model evaluation were implemented in three different settings of 5-fold, Leave One Sample Out (LOSO), and Leave One Participant Out (LOPO) CV [51], [52]. In principle, LOSO should deliver better performance compared to 5-fold and LOPO, as it utilizes more training data in each CV iteration. LOPO, on the other hand, is expected to produce the worst results since, in each CV iteration, the data of the participants being tested is excluded from the training process. However, these approaches are not directly comparable due to differences in the training and test sets.

Table 4 presents the SIS and OHS detection results from the predictive models developed using model-specific selected features for the daily, weekly, and biweekly outcome assignment approaches. In both 5-fold and LOSO CV settings for daily detection, the best performance was achieved using CatBoost as the regression model. The most promising results overall were obtained with CatBoost trained in the LOSO CV setting, leveraging larger training data compared to 5-fold CV. This approach achieved an MAE of 0.75, significantly lower than one unit of SIS. On the other hand, as expected, the results of LOPO are inferior to LOSO and 5-fold CV. This is due to unique sensor data patterns and health outcomes for individual participants. In Table 4, the models trained in the LOSO CV setting for weekly SIS detection consistently outperformed those trained using 5-fold CV, with BoW outperforming the other models. Overall, the results from the daily outcome assignment approach surpassed those of the weekly and biweekly approaches, demonstrating its superior effectiveness in SIS detection.

**Table 4.** Social Isolation Scale and Oxford Hip Score detection results for the daily, weekly, and biweekly outcome assignment approaches. The best results for each outcome assignment approach and cross validation setting are highlighted in bold.

| Outcome [1] | CV [2] | Model [3] | Social Isolation Scale | | Oxford Hip Score | |
|---|---|---|---|---|---|---|
| | | | MAE [4] | $R^2$ [4] | MAE [4] | $R^2$ [4] |
| Daily | 5-fold | DT | 1.63 | 0.45 | 2.85 | 0.75 |
| | | SVR RBF | 1.65 | 0.50 | 2.58 | 0.83 |
| | | LightGBM | 1.42 | 0.65 | 2.35 | 0.86 |
| | | CatBoost | **1.38** | **0.68** | **2.36** | **0.87** |
| | LOSO | DT | 1.47 | 0.59 | 2.80 | 0.73 |
| | | SVR RBF | 1.39 | 0.62 | 2.53 | 0.83 |
| | | LightGBM | 1.30 | 0.70 | 2.34 | 0.87 |
| | | CatBoost | **0.75** | **0.84** | **1.10** | **0.94** |
| | LOPO | CatBoost | 2.58 | -0.18 | 6.58 | -0.21 |
| Weekly | 5-fold | BoW | **1.48** | **0.60** | **3.55** | **0.69** |
| | | LSTM | 2.14 | 0.19 | 3.60 | 0.70 |
| | | TCN | 2.51 | 0.15 | 4.49 | 0.56 |
| | | ROCKET | 2.45 | 0.08 | 5.55 | 0.38 |
| | LOSO | BoW | **1.44** | **0.60** | **3.29** | **0.72** |
| | | LSTM | 2.01 | 0.26 | 3.80 | 0.69 |
| | | TCN | 2.24 | 0.19 | 3.78 | 0.59 |
| | | ROCKET | 1.66 | 0.47 | 3.92 | 0.58 |
| Biweekly | 5-fold | BoW | 1.79 | 0.32 | 3.78 | 0.60 |
| | LOSO | BoW | 0.99 | 0.68 | 2.83 | 0.69 |

[1] Outcome assignment approach, refer to Figure 2.

[2] 5-fold, Leave-One-Sample-Out (LOSO), and Leave-One-Participant-Out (LOPO) Cross Validation (CV). The training was performed on the first 16 most important sensor features and the first 2 most important demographic information selected through the recursive feature elimination approach for each model.

[3] Decision Tree (DT), Support Vector Regressor (SVR) with Radial Basis Function (RBF) kernel, Light Gradient-Boosting Machine (LightGBM), Categorical Boosting (CatBoost), Bag of Words (Bow) followed by CatBoost, Long Short-Term Memory (LSTM), Temporal Convolutional Network (TCN), and ROCKET.

[4] Mean Absolute Error (MAE) and coefficient of determination (R2) between the outcome measurements by the models and the ground-truth outcomes.

## 5. Usage Notes
### 5.1. Dataset Potential Applications

The MAISON-LLF Dataset supports supervised machine-learning and deep-learning model development, using multimodal sensor data as input to estimate health outcomes such as SIS, OHS, OKS, TUG, and the 30-second chair stand test. With sensor data and corresponding ground-truth outcomes spanning multiple weeks, models can be trained for detection (estimating outcomes at the same timestamps as the sensor data) or prediction (estimating future outcomes based on the sensor data). The dataset includes not only the final clinical scores, which are the summation of individual items, but also the values of the individual items themselves. This enables the development of item-specific models or multi-label models. The dataset also provides opportunities to explore model personalization, enabling the development of tailored machine-learning models that account for individual differences in sensor data patterns and health outcomes.

The dataset is also suitable for unsupervised machine-learning approaches, including multimodal or spatio-temporal clustering, to identify patterns within the sensor data without relying on clinical outcomes. Additionally, it enables the study of correlations and associations among sensor modalities, clinical outcomes, and the interplay between them.

Furthermore, by offering a publicly available dataset that specifically focuses on older adults, the MAISON-LLF dataset helps address the lack of accessible data representing marginalized groups, supporting more inclusive machine-learning research [59]. This approach ensures that older adults' unique needs and experiences are recognized and integrated into model development, contributing to the mitigation of digital ageism in AI [60].

## 5.2. Dataset Limitations

Although the dataset is novel and unique in its population, study settings, and types of data collected, it has certain limitations. The relatively small sample size restricts the generalizability of the models trained on it to a broader population. The data collection for MAISON-LLF is ongoing and aims to increase the number of participants to 20. Additionally, the dataset is limited to urban-dwelling older adults living alone in the Greater Toronto Area, Canada, restricting the applicability of the models developed using it to those in rural or remote areas or those living with others. Finally, the supervised models built with the dataset rely on objective data, such as sensor data and demographics, while the ground-truth labels used during training are derived from subjective questionnaires (e.g., SIS and OHS), potentially introducing a mismatch that affects the accuracy and reliability of the models' predictions.

## 6. Code Availability

The dataset is publicly accessible on Zenodo [50], and the code for data preprocessing, feature extraction, feature selection, and supervised model development is hosted on GitHub [61]. The code is written in Python, primarily utilizing libraries such as Pandas, NumPy, Scikit-learn, and PyTorch.

**Acknowledgements**

The authors thank Faranak Dayyani for her invaluable assistance with sensor installation and removal, as well as her dedicated work in biweekly clinical data collection. This work was supported by Shehroz S. Khan and Charlene H. Chu funding from Centre for Aging + Brain Health Innovation (CABHI) Grants, as well as Shehroz S. Khan and Charlene H. Chu's Natural Sciences and Engineering Research Council of Canada (NSERC) Discovery Grants.

**Author Contributions**

Shehroz S. Khan envisaged the MAISON platform. Ali Abedi designed and developed the MAISON platform, including front-end, hardware, and back-end components. Ali Abedi prepared and submitted the required documents to the research ethics board to facilitate dataset publication, with support from Charlene H. Chu and Shehroz S. Khan. Ali Abedi conducted data preprocessing, feature extraction, predictive modeling, and interpretation. Additionally, Ali Abedi drafted the manuscript and all sections were reviewed and finalized by all authors. Charlene H. Chu and Shehroz S. Khan were responsible for designing the clinical and technical aspects of the project. They also provided comprehensive support for its execution.

**Competing Interests**

The authors declare no competing interests.

[56] G. Ke *et al.*, "Lightgbm: A highly efficient gradient boosting decision tree," *Advances in neural information processing systems*, vol. 30, 2017.
[57] A. Abedi, C. Thomas, D. B. Jayagopi, and S. S. Khan, "Bag of states: a non-sequential approach to video-based engagement measurement," *Multimedia Systems*, vol. 30, no. 1, pp. 1–16, 2024.
[58] A. Dempster, F. Petitjean, and G. I. Webb, "ROCKET: exceptionally fast and accurate time series classification using random convolutional kernels," *Data Mining and Knowledge Discovery*, vol. 34, no. 5, pp. 1454–1495, 2020.
[59] C. H. Chu *et al.*, "Digital ageism: challenges and opportunities in artificial intelligence for older adults," *The Gerontologist*, vol. 62, no. 7, pp. 947–955, 2022.
[60] C. Chu *et al.*, "Strategies to Mitigate Age-Related Bias in Machine Learning: Scoping Review," *JMIR aging*, vol. 7, p. e53564, 2024.
[61] A. Abedi, "Codes for MAISON-LLF: Multimodal AI-Based Sensor Platform for Older iNdividuals-Lower Limb Fracture," 2025. [Online]. Available: https://github.com/abedidev/maison-llf